\documentclass[journal]{IEEEtran}

\ifCLASSINFOpdf
  \usepackage[pdftex]{graphicx}
  \graphicspath{{../pdf/}{../jpeg/}}
  \DeclareGraphicsExtensions{.pdf,.jpeg,.png}
\else
\fi
\usepackage{amsmath}
\usepackage{amsfonts}
\usepackage[ruled]{algorithm2e}
\usepackage{algorithmic}

\usepackage{subcaption,xcolor}

\begin{document}
\title{Failure-tolerant Distributed Learning for Anomaly Detection in Wireless Networks}

\author{Marc~Katzef, %~\IEEEmembership{Member,~IEEE,}
        Andrew~C.~Cullen, %~\IEEEmembership{Member,~IEEE,}
        Tansu~Alpcan,~\IEEEmembership{Senior~Member,~IEEE,}
        Christopher~Leckie, %~\IEEEmembership{Member,~IEEE,}
        and~Justin~Kopacz%~\IEEEmembership{Member,~IEEE,}
\thanks{Marc~Katzef and Tansu~Alpcan are with the Department of Electrical and Electronic Engineering, The University of Melbourne, Australia.}%
\thanks{Andrew~C.~Cullen and Christopher~Leckie are with the School of Computing and Information Systems,
The University of Melbourne, Australia.}%
\thanks{Justin~Kopacz is with the Northrop Grumman Corporation, USA}
\thanks{Manuscript received Month Day, Year; revised Month Day, Year.}}

% The paper headers
\markboth{}%
{Katzef \MakeLowercase{\textit{et al.}}: Failure-tolerant Distributed Learning for Anomaly Detection in Wireless Networks}
% The only time the second header will appear is for the odd numbered pages
% after the title page when using the twoside option.
%
% *** (!) Note that you probably will NOT want to include the author's ***
% *** name in the headers of peer review papers.                   ***
% You can use \ifCLASSOPTIONpeerreview for conditional compilation here if
% you desire.

\maketitle

\begin{abstract}
The analysis of distributed techniques is often focused upon their efficiency, without considering their robustness (or lack thereof).
Such a consideration is particularly important when devices or central servers can fail, which can potentially cripple distributed systems.
When such failures arise in wireless communications networks, important services that they use/provide (like anomaly detection) can be left inoperable and can result in a cascade of security problems.
In this paper, we present a novel method to address these risks by combining both flat- and star-topologies, combining the performance and reliability benefits of both.
We refer to this method as ``Tol-FL", due to its increased failure-tolerance as compared to the technique of Federated Learning.
Our approach both limits device failure risks while outperforming prior methods by up to $8\%$ in terms of anomaly detection AUROC in a range of realistic settings that consider client as well as \textit{server} failure, all while reducing communication costs.
This performance demonstrates that Tol-FL is a highly suitable method for distributed model training for anomaly detection, especially in the domain of wireless networks.

\end{abstract}

% Note that keywords are not normally used for peerreview papers.

% For peer review papers, you can put extra information on the cover
% page as needed:
% \ifCLASSOPTIONpeerreview
% \begin{center} \bfseries EDICS Category: 3-BBND \end{center}
% \fi
%
% For peerreview papers, this IEEEtran command inserts a page break and
% creates the second title. It will be ignored for other modes.
\IEEEpeerreviewmaketitle

\section{Introduction}
% Distributed ML for security
% \IEEEPARstart{A}{chieving} wireless network security requires significant effort, mitigating the risks of system breaches. This topic has earned a significant research effort that has been devoted to anomaly detection to improve security, with new methods presented in lockstep with a changing technological landscape.
\IEEEPARstart{I}{mproving} wireless network security through anomaly detection has attracted a significant research effort, with new methods presented in lockstep with a changing technological landscape.
As this landscape has evolved to include networks with an increasing number of devices, approaches to detect anomalies have become more reliant on automated methods offering pattern recognition.
Machine Learning (ML) has found use in rapidly-evolving, critical applications like distributed network security because of its flexibility and ability to process high dimensional data \cite{zhang_deep_2018, liu_survey_2017}.

ML-based anomaly detection models can identify if unusual (and possibly harmful) activity is present in an observed network.
Observations of a wireless network may be produced by devices scattered over a large spatial area, producing partially overlapping observations.
This data is readily available at each wireless device, along with storage and computational resources.
However, in many contexts, the data cannot be easily centralised for use in ML training \cite{liu_survey_2017}.
Recognising the risks and resource usage imposed by transferring data from distributed devices to a centralised point, recent research has focused upon the development of schemes in which locally-trained models are exchanged with their peers \cite{jiang_collaborative_2017}
 One popular example of such edge computing is Federated Learning (FL) \cite{mcmahan2017communication}, which combines distributed computation with centralised aggregation, thus removing the need for exchanging datasets \cite{mcmahan2017communication}. 

\begin{figure}[h]
	\centerline{\includegraphics[width=0.45\textwidth]{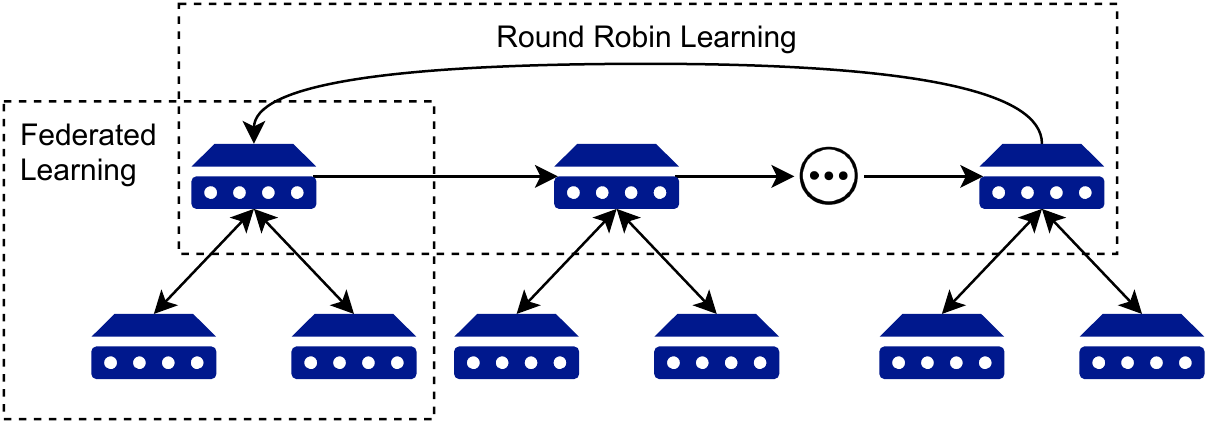}}
	\caption{Star and flat topologies can coexist as parts of a single network, as indicated by dashed borders. Each cluster (star topology) comprises a server and multiple clients, and the flat topology comprises the servers of the star topologies in the network.}
	\label{fig:tol_fl_top}
\end{figure}

A single round of FL's core algorithm consists of selecting a subset of clients, sending to them the most recent model parameters, and allowing these clients to further train the model on local data before averaging their locally-updated models.
While such a scheme has significant advantages in terms of both parallel processing and data privacy, it has an oft-overlooked weakness: that its inherent star topology introduces systemic fragility, due to the risk of failure of the centralised server. 
A common assumption with prior distributed ML works is that the server is highly reliable and that the main failure or attack modes are through the client devices instead \cite{mcmahan2017communication, bonawitz_practical_2017, dinh_federated_2021}.
This hardware dependence is highlighted by Figure~\ref{fig:tol_fl_top}, in which there is a clear distinction between star- and flat-topology training, with star-topologies depending on a central device.
Due to FL's inherent need for a server, the robustness of such a training scheme to device failures has received little attention.
This is a significant omission, as critical infrastructure should always be considered in context of the question \textit{what happens when devices fail}?
Such a consideration is especially important in distributed training due to the large set of potential failure modes, including those relating to connectivity, device behaviour, and resources or bandwidth. 
While distributed training schemes have the potential to
\textit{continue} training in the event of device failure \cite{kwon_survey_2019, zhang_deep_2018}, a level of robustness and redundancy is required in the training scheme to enable the remaining devices to seamlessly resume training where a faulty/compromised device failed.
Such robustness is especially important for critical applications, such as those involving network security.

Prior works have suggested building on FL with extensions that improve training speed, such as FedGroup \cite{duan_fedgroup_2021}, IFCA \cite{NEURIPS2020_e32cc80b}, and FeSEM \cite{xie_multi-center_2021} (all of which train multiple models in parallel in the network).
To the authors' knowledge, however, this paper is the first to apply a clustered training scheme consisting of both FL and a flat-topology, round robin learning scheme to the wireless network anomaly detection field.
In this work, we propose a novel machine learning training scheme that was designed specifically for wireless network applications, and analyse the impact of client \textit{or server} failure.

This paper addresses a distributed training research gap of strengthening such methods for the case of crucial devices failing while training an anomaly detection model.
We address this problem by first presenting an algorithm (Tol-FL) along with a theoretical analysis of its ability to continue training after any one device is removed.
After establishing this method, Tol-FL, we compare the performance of this training algorithm with existing methods on the task of anomaly detection when exposed to a simulated network scenario that contains examples of resource misuse and novel network traffic patterns as well as random device failures, representing a range of realistic settings.
The contributions of this paper are:
\begin{enumerate}
	\item A novel ML training scheme---we call Tol-FL---that combines hierarchical and flat network topologies
	\item A new parametrisation, $k$ that allows for Tol-FL to smoothly transition between flat and hierarchical forms---balancing the performance benefits of each
	\item A complexity analysis of the failure tolerance and convergence speed of FL variants in non-ideal networks in which devices or communication links may fail throughout training.
\end{enumerate}

\section{The Problem of Failure-Tolerant Distributed Anomaly Detection}
% Distributed training demands specialised training schemes
% When the thing being trained is for security, additional measures are needed
% Robustness in this paper is about tolerance to device failure

Centralising data collected by distributed devices inherently presents both privacy and security risks, while also introducing communication costs. To circumvent these limitations, a range of specialised ML variants have been proposed, in which models are trained over distributed devices that have access to only parts of the full dataset.
While some candidate training schemes like FL appeared in literature \cite{mcmahan2017communication}, they bring new risks and challenges that must be addressed.
One major risk is the presence of a central server---typically a critical component for the training algorithm to operate.
The risk that the central server poses is one of reliability, wherein, a server failure brings any ongoing tasks to a halt.
With ML being increasingly used in tasks like network anomaly detection, such an outage can affect the security of remaining devices.

In this paper, we consider the case where individual devices or a server that is participating in a distributed training scheme may be brought offline at random, either due to a targeted attack, or simply due to environmental conditions.
Through such an outage, any \textbf{one networked device may become unreachable} at any stage during network operation and---our main focus---\textbf{during the training phase} of an anomaly detection system.
For star-based topologies, if the failing entity is a server, this event can prevent communication amongst all the remaining devices.
This worst-case scenario separates the remaining devices, their computational resources, and their datasets.
With the possibility of client \textbf{or server} failure, we seek to maximise the performance of the resulting model without sacrificing performance in other scenarios or increasing resource usage.

% ====================
\section{Failure Tolerant Federated Learning}
%Plan:
%Outline
%	Motivation
%	Main components
%	Novelty
%Structure
%Sequence diagram
%Algorithm
%Equations needed (including fedavg and optionally SBT)
%Summary - expectations about performance

Tol-FL has been designed to reduce the dependence on a single device that is found in star-topology schemes by combining FL with an entirely flat topology training scheme. 
Tol-FL is the combination of two opposing distributed training schemes: FL and a completely flat-topology alternative to FL that has been used as an example in prior works such as \cite{chou_efficient_2021, yang_scheduling_2019}.
To the authors' knowledge, this flat-topology technique has not yet been assigned a name; to simplify further discussion, we refer to this flat, multi-device, round robin-based ML training scheme as Super-Batch Training (SBT).
Through this combination, the system is hardened against the potential failure of individual clients or servers---minimising any degradation to performance or resource usage under such conditions.

% While FL does x, SBT does y
% this behaviour results in the following shortcomings of each
Tol-FL has been designed to balance the benefits and drawbacks inherent in
star and flat topologies.
While FL and other star-topology based schemes can offer high levels of parallel processing (speeding up training time), this scheme comes with the cost of relying on a highly connected and dependable server being live without failure.
SBT by comparison, trains a model in sequence by iterating over all of the participating devices' local datasets as if they were mini-batches on a server, multiple times (further detailed in Appendix~\ref{app:sbt}).
This flat-topology training approach divides the tasks of a central server over many non-critical devices, increasing training robustness.
However, this improved training robustness comes at the cost of an increased number of sequential events (rather than parallel) and thus an increase in training time.

% Tol-FL combines these two techniques as shown in
% In this structure, we have FL within clusters and SBT over clusters
% The processing happening within each group is fedavg
% The processing happening over each group is SBT
% This is shown more formally in algorithm ...
Tol-FL addresses the respective shortcomings of SBT and FL by combining the two algorithms to provide the structure in Figure~\ref{fig:tol_fl_top}.
Instead of using a flat topology \textit{or} a star topology exclusively, Tol-FL is the composition of FL and SBT in which FL is performed within subgroups of devices in the network (using an arbitrary member device as the server) and SBT is applied over these subgroups.
Through this combination, Tol-FL benefits from FL's training speed and SBT's tolerance to server/device failure.
In this combined structure, we introduce a new design parameter---the number of clusters $k$ in the network---which allows us to smoothly transition between FL and SBT to provide a required balance of robustness and speed.
This parameter and other required terms in Table~\ref{tbl:notation} appear in the formal definition of Tol-FL given in Algorithm~\ref{alg:superbatch-fl-hybrid}.

\begin{algorithm}[tb]
   \caption{One round of Tol-FL training for a network containing $N$ devices, $D = \{d_i\}_{i=1}^{N}$, each assigned to one of $k$ clusters $D_i, i \in \{1..k\}$ that are non-overlapping ($D_i \cap D_j = \emptyset \forall i \neq j$ and $\bigcup_{i=1}^{k}D_i = D$).}
   \label{alg:superbatch-fl-hybrid}
	\begin{algorithmic}
	\REQUIRE $D, k, \theta_t$ // Devices, number of clusters, current model parameters
	\STATE // In each cluster, \textbf{in parallel}:
	\FOR {i = 1, 2, ..., $k$}
		\STATE $n_{t, i}, g_{t, i} = \text{FedAvg}(D_i, \theta_t)$
	\ENDFOR

	// Initialise on the first cluster head in a sequence
	\STATE $g_{t} = \textbf{0}$ // Mean gradient
	\STATE $n_{t} = 0$  // Sample count
	\STATE // In each cluster, in sequence:
	\FOR {i = 1, 2, ..., $k$}
		\STATE $n_{t} \leftarrow n_{t} + n_{t, i}$  // Record the number of samples in cluster
		\STATE $r \leftarrow \frac{n_{t, i}}{n_{t}}$  // Calculate a weighting factor
		\STATE $g_{t} \leftarrow r g_{t, i} + (1 - r) g_{t} $ // Update the mean gradient
	\ENDFOR
	\STATE // On the last cluster head of the sequence, apply gradient update
	\STATE $\theta_{t+1} = \theta_{t} - \alpha g_{t}$
	\STATE // Broadcast the latest parameters $\theta_{t+1}$
	\end{algorithmic}
\end{algorithm}

The two main steps in Tol-FL consist of a parallel processing stage in each of $k$ clusters, followed by sequential aggregation of each cluster's output.
These two steps appear in Algorithm~\ref{alg:superbatch-fl-hybrid} as two iterative processes; one involving FL's FedAvg algorithm, and the other using the SBT algorithm.
In each round of Tol-FL training, every device optimises a copy of the global model on their local dataset before passing their updated model to an elected device (the cluster head).
The cluster head aggregates all the updated models from devices in its cluster using FedAvg to determine $g_{t, i}$ and $n_{t, i}$ (the per-cluster gradient and number of samples used) before transferring the aggregated model \textit{along with the number of samples used in this update} to a neighbouring cluster head.
These cluster head messages can be combined easily in sequence using SBT, before finally broadcasting the updated model to all participating devices.
Each iteration of the model is generated using the well-known \cite{goodfellow_deep_2016} model update form 
\begin{equation*}
\theta_{t+1} = \theta_{t} - \alpha g_{t}
\end{equation*}
however, to obtain $g_t$, our iterative update method is used over clusters $i$ through $k$ to combine their cluster head messages as
\begin{equation*}
g_t = \sum_{i=1}^{k} \left(
r_i g_{t, i} + \left(1 - r_i\right) \sum_{j=1}^{i-1}g_{t, j}
\right)
\end{equation*}
where $r_i = {n_{t, i}}/{\sum_{j=1}^{i}n_{t, j}}$ weights each gradient by the number of samples used to calculate it.

\begin{figure}[h]
	\centerline{\includegraphics[width=0.3\textwidth, trim={0 0 0 1cm},clip]{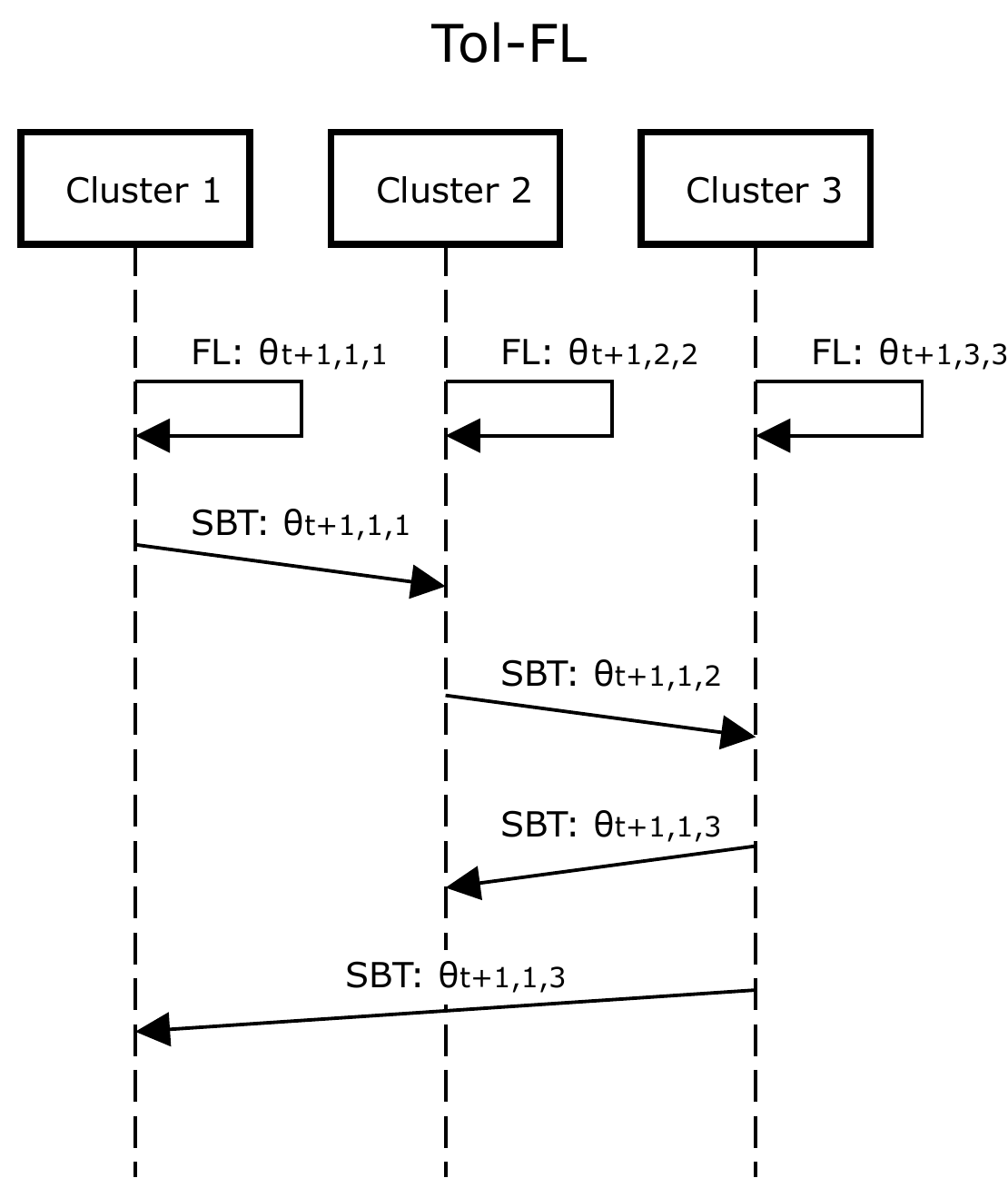}}
	\caption{Tol-FL sequence diagram showing all communication steps involved in one global model update, where each cluster is initialised with $\theta_{t,1,k}$, and updates it to $\theta_{t+1,1,k}$ (the next iteration that considers training sets from clusters $1$ through $k$)}
	\label{fig:tolfl_seq}
\end{figure}

This iterative training is performed as per the sequence shown in Figure~\ref{fig:tolfl_seq}.
A notable property of this sequence is how it varies with the number of clusters, $k$; as $k$ varies, Tol-FL's behaviour falls into one of three categories.
With $k=1$, all devices belong to the same cluster, and so FL's FedAvg algorithm is performed over the entire network---this results in standard FL with high parallelisation but low robustness due to the cluster head device taking on the role of an FL server.
With $k=N$, all clusters consist of just a single device and as a result, Tol-FL is reduced to standard SBT.
In the remaining case of $1 < k < N$, we see a balance between parallel processing and robustness.
Notably, in every value for $k$, the global model is updated with respect to every data sample in the network (with equal weighting).
By construction, this property of Tol-FL means that model updates from a round of training \textit{are independent of} $k$ and result in identical outputs (ignoring low levels of numerical error, which would arise in practice).
With the model output left unchanged, the parameter $k$ allows us to change the division of tasks amongst devices and profoundly change the expected reliability, communication costs, and processing times.

\begin{table}
\centering
\caption{Collaborative training scheme notation.}
\label{tbl:notation}
\begin{tabular}{c|l}
\hline
\hline
\textbf{Symbol} & \textbf{Meaning} \\
\hline
$N$ & Total number of devices in distributed case \\
$N_i$ & Number of devices in cluster $i$ \\
$k$ & Number of clusters \\
$n_{t,i}$ & Number of samples at time $t$, device $i$ \\
$\textbf{x}$ & Data sample \\
$\hat{\textbf{x}}$ & Reconstructed data sample \\
$\theta_{t,i}$ & Model parameters at time $t$, device $i$ \\
$E$ & Number of local epochs \\
$\alpha$ & Learning rate \\
$g_{t,i}$ & Gradients at time $t$, device $i$ \\
\hline
\end{tabular}
\end{table}

%%%%%%%%%%%%%%%%%%%%%%%%%%%%%%%%%%%%%%%%%%%%%%%%%%%%%%
\section{Analysis of Tol-FL Robustness and Resource Usage}
In this section we analyse the resource usage of Tol-FL during training and its robustness to random client or server failure.

\subsection{Tol-FL's Training Rate}
\label{sec:training}
As each topology requires a different sequence of communications and processing, the amount of time they need for their training to converge varies.
This time is determined by task sequencing, available data rates, and CPU resources across the set of devices.
To evaluate the resource requirements of each of the competing methods, we present the resource usage of competing techniques in Table~\ref{tbl:resources}.

\begin{table*}
	\centering
	\caption{Resource usage of Tol-FL, SBT and existing methods during training, where $s$ is the number of samples, $d$ is the number of devices, and $k$ is the number of clusters. Note, communication resources are further broken down by sources and destinations with, for example, C2S standing for Client to Server.}
	\label{tbl:resources}

	\begin{tabular}{c|c|c|c|c|c|c}
		\hline
		\hline
		\textbf{Method} & \textbf{Comp. (Server/s)} & \textbf{Comp. (Client/s)} & \textbf{Comms (S2S)} & \textbf{Comms (C2S)} & \textbf{Comms (C2C)} & \textbf{Time} \\
		\hline
		Batch & $\mathcal{O}(s)$ & $-$ & $-$ & $-$ & $-$ & $\mathcal{O}(s)$ \\
		\hline
		FL & $\mathcal{O}(d)$ & $\mathcal{O}(s)$ & $-$ & $\mathcal{O}(d)$ & $-$ & $\mathcal{O}(s + d)$ \\
		\hline
		\textbf{Super-Batch} & $-$ & $\mathcal{O}(s)$ & $-$ & $-$ & $\mathcal{O}(d)$ & $\mathcal{O}(s + d)$ \\
		\hline
		\textbf{Tol-FL} & $\mathcal{O}(d)$ & $\mathcal{O}(s)$ & $\mathcal{O}(k)$ & $\mathcal{O}(d)$ & $-$ & $\mathcal{O}(s + k)$ \\
		\hline
	\end{tabular}
\end{table*}

% (!) possibly add discussion on resource usage

\subsection{Tol-FL's Tolerance to Server Failure}
One of Tol-FL's greatest strengths is its ability to continue training even in the event of an individual device (client or server) failing during training.
In such an event, a single device going offline \textit{cannot} force the entire network to cease training collaboratively---instead, only the subset of devices that depended on the failing device are removed from training.
This new worst-case scenario can be seen as a method of mitigating a training-time risk that would ordinarily be enticing to malicious attackers.

The way in which Tol-FL is able to continue training can be seen through Algorithm~\ref{alg:superbatch-fl-hybrid}.
This algorithm shows that Tol-FL divides devices into clusters and that if one device were to fail throughout the network, one of two scenarios would occur, depending on the device's role inside the cluster to which it has been assigned (i.e., cluster member or cluster head). If the device was a standard cluster member (i.e., \textit{not} the cluster head) then the cluster may continue participating in distributed training as usual, only without the use of the failed device's resources---data and processing.
If, instead, the device was designated as the cluster head---the one member of the cluster responsible for inter-cluster communication---then the entire cluster would be absent from subsequent training.
Subsequently, the worst-case scenario for Tol-FL would be losing a cluster head, which would result in losing access to one entire cluster $D_i$.
The devices in the remaining $k-1$ clusters $D_j, \forall j \neq i$ may continue training collaboratively.

Tol-FL's continued training is in stark contrast to standard FL, in which the following failure modes exist: if a FL client fails, subsequent training can no longer use that device's data or processing resources; if the FL server fails, distributed training is terminated and client devices can no longer benefit from other devices' resources.
From these FL failure scenarios, the worst-case failure mode is again from server failure, leaving the remaining devices with the task to make alternative arrangements such as assigning a new FL server---a non-trivial task given that the FL server is often much more powerful than any one client---or training locally on each client.
Notably, such arrangements were not defined in the original FL formulation and are often overlooked in FL research.
By comparing standard FL with Tol-FL, we see that the worst-case scenario for a device failure is more detrimental to FL. How this translates into real-world performance is explored in Section~\ref{sec:experiments}.

Performance in practice depends on the likelihood of different failure events occurring.
This performance can be approximated from device failure scenarios that are formulated to display the best- and worst-case device failure modes in the compared training methods.
Expected performance (e.g., loss or anomaly detection score, $J$), however, is expressed as $\sum_{s \in S} p_s \cdot J_s$.
Depending on the probability of any client or server failing and the performance each method achieves in that scenario, either batch, FL, or Tol-FL may be most suited to the application.

% ====================
\section{Tol-FL Experimental Evaluation}
\label{sec:experiments}
The following sections present a series of experiments that have been designed to measure the performance of an anomaly detection scheme's training with and without a client or server failure.
These experiments have been designed to mimic realistic scenarios that can result in performance drops due to devices becoming unavailable during training---the real-world performance drops that Tol-FL aims to ameliorate.

% --------------------
\subsection{Experimental Setup and Baseline}
\label{sec:exp}
Tol-FL's performance was measured and compared with existing methods of batch training, FL, FedGroup, IFCA, and FeSEM in multiple scenarios, using multiple datasets which have been assigned to devices for a realistic representation of network security tasks.
The task in question is to train an autoencoder anomaly detection model on a range of datasets that are representative of wireless network environments.
The autoencoder model trained in these experiments consists of a fully-connected artificial neural network for both the encoder and decoder, trained to minimize reconstruction error $J(\textbf{x}) = ||\textbf{x} - \hat{\textbf{x}}||_2^2$. Each model was designed with three hidden layers with $64$ to $128$ neurons per layer and a code vector of length $32$.
ReLU was used as the activation function for hidden layers and linear was used for output.
Following \cite{goodfellow_empirical_2015}, dropout was used during training for all hidden layers using a probability of $0.2$.
When presented with a new sample $\mathbf{x}$, this state-of-the-art method's reconstruction error $J(\mathbf{x})$ indicates how similar this sample is to previous samples, leading to $J(\mathbf{x})$ being used as the anomaly scoring function.
% (!) citation would be useful

The selected anomaly detection model was trained using the above methods (Tol-FL, FL, FedGroup, IFCA, and FeSEM) on the authors' publicly available dataset generator---Comms-ML \cite{katzef_commsml}---as well as a collection of commonly-used reference datasets described in Appendix~\ref{app:datasets}.
Together, these datasets give an accurate representation of wireless networks both in terms of sample structure and in terms of data layout---distributed over multiple devices.
For datasets that are not naturally divided into region/device datasets, the data was subdivided in multiple arrangements based on class to test both centralised and decentralised method performance.
For all decentralised experiments, these data divisions occurred by separating the overall dataset into approximately equally-sized subsets of length $|D_i| = N_i \leq \lceil \frac{N}{k} \rceil$ for each cluster.
Within each cluster, this division method is performed yet again to allocate the sub-dataset (of any one cluster) to devices that belong to that sub-dataset's cluster.

\begin{figure}[h]
	\centerline{\includegraphics[width=0.45\textwidth]{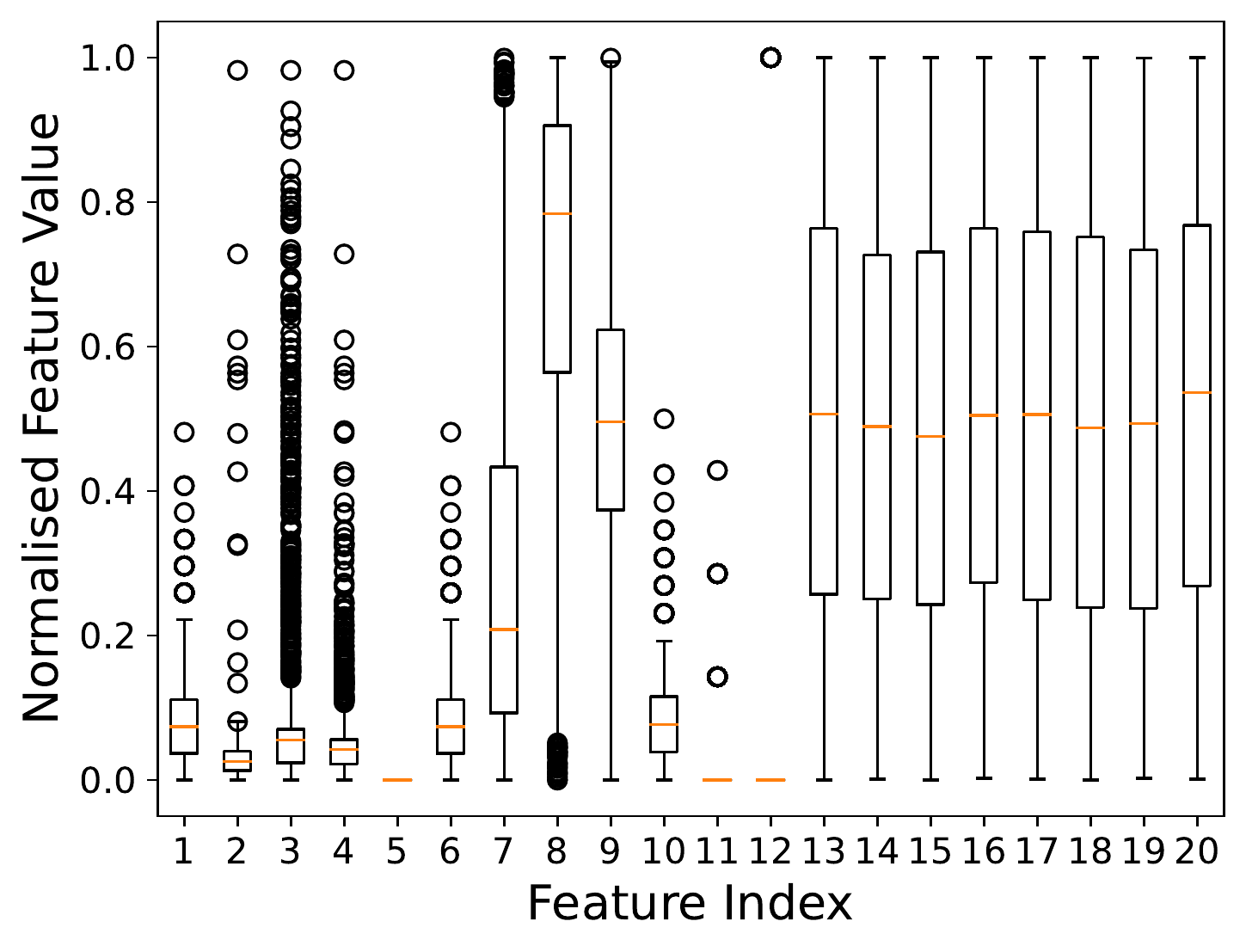}}
	\caption{An excerpt of (20 out of 112 features of) wireless network activity showing the mean values of statistics-based features (indices 0--11) and raw physical signals (indices 11 onward) for one class of the Comms-ML dataset. Successful anomaly detection requires distinct values or patterns to be found between typical and anomalous classes.}
	\label{fig:demo_sample}
\end{figure}

Using Comms-ML, dividing a dataset across multiple devices is easily achieved by defining custom networks, giving a flexible way to investigate the scenario of wireless communications.
Generated through simulations, the Comms-ML dataset contains raw in-phase/quadrature data as well as network statistics that would be available to devices in an SDR network in which multiple devices were simulated to communicate in multiple, distinct geographic regions using Wi-Fi.
The communications within these networks are recorded in two formats---a compact transcript of the transmitted data and a realistic, wide-band waveform received from each of the devices in the region.
These data types both appear in each sample as either raw readings or signal statistics as in Figure~\ref{fig:demo_sample}.

Using this wireless signal generator, anomalies have been introduced into the network by adjusting the communication patterns of existing devices (e.g., more or less frequent transmissions) or adding novel devices to the environment (e.g., Bluetooth devices entering a previously WLAN-only network).
Each of these anomaly types changes the underlying data distribution in different ways, and should be identified by a well-trained anomaly detector.
Table~\ref{tbl:benchmark_clean} confirms that all competing training methods achieve strong performance across all datasets when they are free to train without any server/device failure occurring.
While most results are clustered, FL and Tol-FL are seen to be the top two methods in this case, showing promise for use in ideal, failure-free scenarios---scenarios that we will use as a baseline.
In the next two sections, we explore the scenarios in which client or server failures occur partway through training.

\begin{table*}
\centering
\caption{
Anomaly detection AUROC results \textbf{without server or device failure}.
Here, $^\dagger$ denotes schemes that train multiple instances of the same model and $^*$ denotes the top-performing instance from such schemes.
}
\label{tbl:benchmark_clean}
\begin{tabular}{c|ccccc|cccc}
\hline
\textbf{Dataset} & \textbf{Tol-FL} & FedGroup$^*$ & IFCA$^*$ & FeSEM$^*$ & FL & Batch & FedGroup$^\dagger$ & IFCA$^\dagger$ & FeSEM$^\dagger$ \\
\hline
\textbf{Comms-ML} & $\mathbf{0.80 \pm 0.01}$ & $0.75 \pm 0.02$ & $0.76 \pm 0.02$ & $0.79 \pm 0.01$ & $\mathbf{0.80 \pm 0.01}$ & $0.82 \pm 0.01$ & $0.84 \pm 0.02$ & $0.85 \pm 0.03$ & $0.82 \pm 0.02$ \\
\textbf{FMNIST} & $0.95 \pm 0.02$ & $0.88 \pm 0.04$ & $0.89 \pm 0.03$ & $0.92 \pm 0.03$ & $\mathbf{0.96 \pm 0.02}$ & $0.96 \pm 0.02$ & $0.97 \pm 0.02$ & $0.97 \pm 0.02$ & $0.96 \pm 0.02$ \\
\textbf{CIFAR-10} & $0.71 \pm 0.05$ & $0.69 \pm 0.04$ & $0.70 \pm 0.05$ & $0.71 \pm 0.04$ & $\mathbf{0.73 \pm 0.04}$ & $0.72 \pm 0.04$ & $0.74 \pm 0.03$ & $0.74 \pm 0.04$ & $0.73 \pm 0.05$ \\
\textbf{CIFAR-100} & $0.75 \pm 0.04$ & $0.73 \pm 0.04$ & $0.72 \pm 0.04$ & $0.73 \pm 0.05$ & $\mathbf{0.77 \pm 0.05}$ & $0.78 \pm 0.03$ & $0.80 \pm 0.02$ & $0.77 \pm 0.03$ & $0.76 \pm 0.03$ \\
\hline
\end{tabular}
\end{table*}

% --------------------
%\subsection{Failure Tolerance Comparison}
%\label{sec:benchmark_ffree}
%The severity of a device failure depends on the roles assigned to that device.
%The distributed schemes we compare in this paper consist of two roles; clients and (optionally) servers. 
%As all practical devices are susceptible to failure, these two roles are considered in our analysis of failure tolerance in the following two subsections. In these subsections, performance is measured with different device failures, and compared against the failure-free scenario.

\subsection{Client Failure}
Client failures have been considered in multiple past works, including the paper introducing FL \cite{mcmahan2017communication}.
This failure scenario consists of the server continuing with training as usual, in the absence of any lost clients.
In this scenario, clients may be ignored as they are plentiful when compared with the solitary server and so the performance degradation is estimated to be low; losing only the computational power and dataset of a single device out of the entire network.

The failure of a client represents the `best case' device failure outcome; in other words, should a device failure occur, the loss of a client device is the preferred mode of failure, and is likely to have the smallest impact on the training in terms of both time and performance.
To evaluate this scenario, the failure-free reference experiments described in Section \ref{sec:exp} were repeated with \textit{client} failures introduced at specified times in training to illustrate the impact in performance that any one client can have in the event of failure/harm that prevents the device from participating in training.

Client failures were introduced into our experiments by eliminating a device from training past a certain epoch---the same device at the same epoch in training for each method.
In doing so, the same subset of data and computational resources are removed from the network at the midpoint of training.
As each client contributes a subset of training data, we expect the loss of one client to increase the test loss, which contains elements from \textit{all} classes (including any present on the failed device).
This increase in loss occurs for the remaining devices' model as determined by the importance of the client's data; if the client held unique data that is significant for the testing phase, the network will use the remaining epochs to train on only a subset of the necessary data to arrive at the failure-free loss.
In our client failure experiments (Table~\ref{tbl:client_failure}), this degradation in training losses results in consistently lower performance than the baseline presented in Section~\ref{sec:exp}.

A series of experiments were run for $10$ repetitions each, taking the performance of the global model or (in the case of FedGroup, IFCA, and FeSEM) the best-performing model throughout the network.
The results of these client failure experiments show that the trends in training losses after a \textit{client} device failure are similar across training methods.
These results demonstrate that either FL or Tol-FL always outperforms the remaining methods, likely due to the fact that these methods train an \textit{individual} model for a shared task (representing all data classes) whereas other methods specialise their models to perform better on a subset of data (one cluster's data).
For wireless anomaly detection, such specialisation can be seen as overfitting to that cluster's data, resulting in a higher false positive rate on data from other clusters.

% Note: lower values than clean, and higher values than server failure
\begin{table*}
\centering
\caption{
Anomaly detection AUROC results \textbf{with client failure} occurring after epoch $50$.
Here, $^\dagger$ denotes schemes that train multiple instances of the same model and $^*$ denotes the top-performing instance from such schemes.
}
\label{tbl:client_failure}
\begin{tabular}{c|ccccc|cccc}
\hline
\textbf{Dataset} & \textbf{Tol-FL} & FedGroup$^*$ & IFCA$^*$ & FeSEM$^*$ & FL & Batch & FedGroup$^\dagger$ & IFCA$^\dagger$ & FeSEM$^\dagger$ \\
\hline
\textbf{Comms-ML} & $0.78 \pm 0.02$ & $0.74 \pm 0.03$ & $0.75 \pm 0.02$ & $0.78 \pm 0.02$ & $\mathbf{0.79 \pm 0.02}$ & $0.80 \pm 0.02$ & $0.81 \pm 0.02$ & $0.82 \pm 0.02$ & $0.80 \pm 0.02$ \\
\textbf{FMNIST} & $0.92 \pm 0.02$ & $0.86 \pm 0.03$ & $0.87 \pm 0.04$ & $0.90 \pm 0.03$ & $\mathbf{0.93 \pm 0.03}$ & $0.93 \pm 0.03$ & $0.94 \pm 0.03$ & $0.94 \pm 0.02$ & $0.93 \pm 0.03$ \\
\textbf{CIFAR-10} & $\mathbf{0.71 \pm 0.04}$ & $0.68 \pm 0.03$ & $0.69 \pm 0.05$ & $0.69 \pm 0.05$ & $\mathbf{0.71 \pm 0.05}$ & $0.71 \pm 0.05$ & $0.72 \pm 0.04$ & $0.72 \pm 0.04$ & $0.71 \pm 0.04$ \\
\textbf{CIFAR-100} & $0.73 \pm 0.05$ & $0.71 \pm 0.04$ & $0.71 \pm 0.05$ & $0.72 \pm 0.04$ & $\mathbf{0.75 \pm 0.04}$ & $0.76 \pm 0.05$ & $0.79 \pm 0.03$ & $0.76 \pm 0.04$ & $0.74 \pm 0.04$ \\
\hline
\end{tabular}
\end{table*}

\subsection{Server Failure}
\begin{figure}[h]
	\centerline{\includegraphics[width=0.4\textwidth]{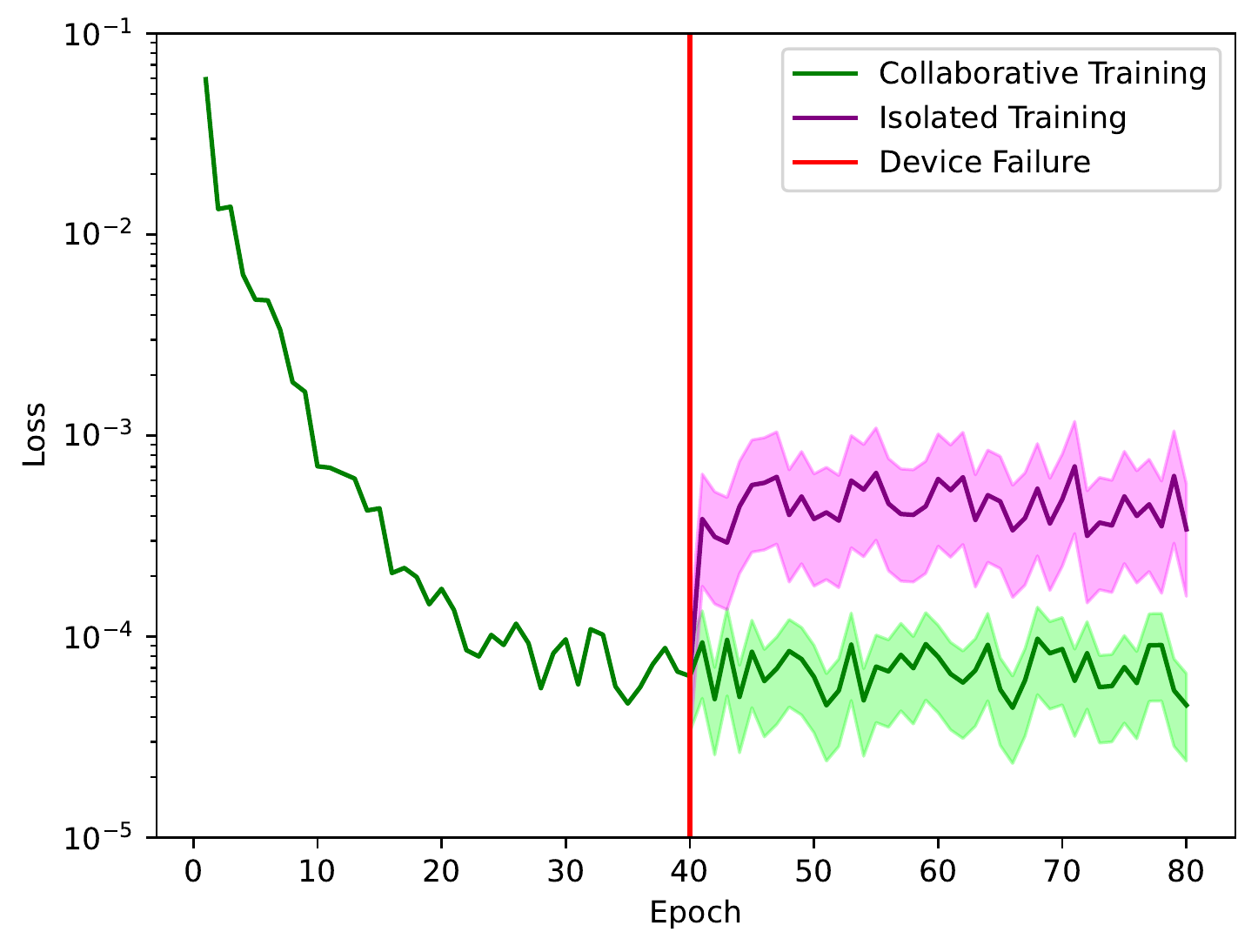}}
	\caption{
Worst-case training result for FL and SBT---isolated or collaborative training, respectively.
These configurations are special cases of Tol-FL: FL is Tol-FL with $k=1$, SBT is Tol-FL with $k=N$.
Results shown are the average of the remaining $N=9$ devices' independent training.
	}
	\label{fig:fl_server_failure}
\end{figure}

Through Section~\ref{sec:training} we have shown that under normal operating conditions, batch training, FL, and Tol-FL converge to optimal solutions (supported by Table~\ref{tbl:benchmark_clean}) and that FL and Tol-FL achieve this after using the same computational budget.
This section explores the largest deviation from normal operating conditions---the case of server failure.
The key difference between FL and SBT---which are Tol-FL with $k=1$ and $k=N$, respectively---is succinctly shown through the cases of server and device failure when training on the MNIST dataset in Figure~\ref{fig:fl_server_failure}.

Compared to client failure, server failure has seen less detailed analysis due to the fact that in many scenarios, the server is essential.
In a fixed, star topology, loss of the server does not merely \textit{reduce} training performance, it terminates all collaborative training (leaving only the option of local training).
It is this restriction that Tol-FL addresses through its hierarchical, distributed structure.
Instead of simply terminating training, Tol-FL can endure any one device failure including the ``servers"/cluster heads of each cluster.
This section demonstrates the impact this tolerance has on training performance.

To each method's anomaly detection performance in the event of a server-failure, the reference experiments detailed in Section~\ref{sec:exp} were repeated with a single server/cluster head failure occurring at a predetermined epoch.
For batch training (which is a centralised training method), such a failure results in the model's training loss plateauing at the final value reached before failure.
For Tol-FL, this failure results in the worst-case Tol-FL performance, in which an entire cluster is removed from training.
For FL, the remaining devices are left with a decision: to train independently on their own datasets or to cease training and keep the final performance in the same way as batch training.

The results of server failure experiments (in Table~\ref{tbl:server_failure}) consistently place Tol-FL's anomaly detection performance above the competitors.
This performance in the event of server failure is due to Tol-FL's distributed structure, in which no single server is essential to proper performance.
Instead, Tol-FL's worst-case performance occurs when a single cluster head device---a device performing FL locally with the members of that cluster---is removed from training.
In this event, the members of that cluster are removed from training, but all other clusters can continue unaffected.
Compared to previous failure modes, the loss of a large portion of data explains why performance is lowest for server failures.
Comparing between methods in this failure mode, the lack of a central server explains why Tol-FL is the least affected and achieves the highest performance.
The performance improvement is most noticeable with the Fashion MNIST dataset, in which Tol-FL outperforms its \textit{closest competitors} by $8\%$.

\begin{table*}
\centering
\caption{
Anomaly detection AUROC results \textbf{with server failure} occurring after epoch $50$.
Here, $^\dagger$ denotes schemes that train multiple instances of the same model and $^*$ denotes the top-performing instance from such schemes.
}
\label{tbl:server_failure}
\begin{tabular}{c|ccccc|ccc}
\hline
\textbf{Dataset} & \textbf{Tol-FL} & FedGroup$^*$ & IFCA$^*$ & FeSEM$^*$ & FL & FedGroup$^\dagger$ & IFCA$^\dagger$ & FeSEM$^\dagger$ \\
\hline
\textbf{Comms-ML} & $\mathbf{0.75 \pm 0.03}$ & $0.71 \pm 0.02$ & $0.70 \pm 0.04$ & $0.71 \pm 0.03$ & $0.64 \pm 0.06$ & $0.75 \pm 0.02$ & $0.76 \pm 0.03$ & $0.80 \pm 0.03$ \\
\textbf{Fashion MNIST} & $\mathbf{0.85 \pm 0.03}$ & $0.79 \pm 0.04$ & $0.79 \pm 0.03$ & $0.78 \pm 0.04$ & $0.65 \pm 0.05$ & $0.88 \pm 0.02$ & $0.91 \pm 0.01$ & $0.89 \pm 0.02$ \\
\textbf{CIFAR-10} & $\mathbf{0.68 \pm 0.04}$ & $0.65 \pm 0.05$ & $0.66 \pm 0.05$ & $\mathbf{0.68 \pm 0.04}$ & $0.59 \pm 0.06$ & $0.70 \pm 0.03$ & $0.71 \pm 0.03$ & $0.70 \pm 0.03$ \\
\textbf{CIFAR-100} & $\mathbf{0.73 \pm 0.02}$ & $0.70 \pm 0.03$ & $0.68 \pm 0.04$ & $0.70 \pm 0.04$ & $0.67 \pm 0.05$ & $0.74 \pm 0.03$ & $0.72 \pm 0.04$ & $0.73 \pm 0.03$ \\
\hline
\end{tabular}
\end{table*}

\subsection{Resource Usage}
While the previous sections show Tol-FL matches or exceeds performance of prior works in different failure cases, any additional functionality like failure tolerance must not only offer strong performance outright, but also suitably low resource usage in order to be practical.
To verify that Tol-FL meets this requirement, we measure processing time and communication usage---two critical factors to consider in a wireless device's operation.

\begin{figure}[h]
	\centerline{\includegraphics[width=0.45\textwidth]{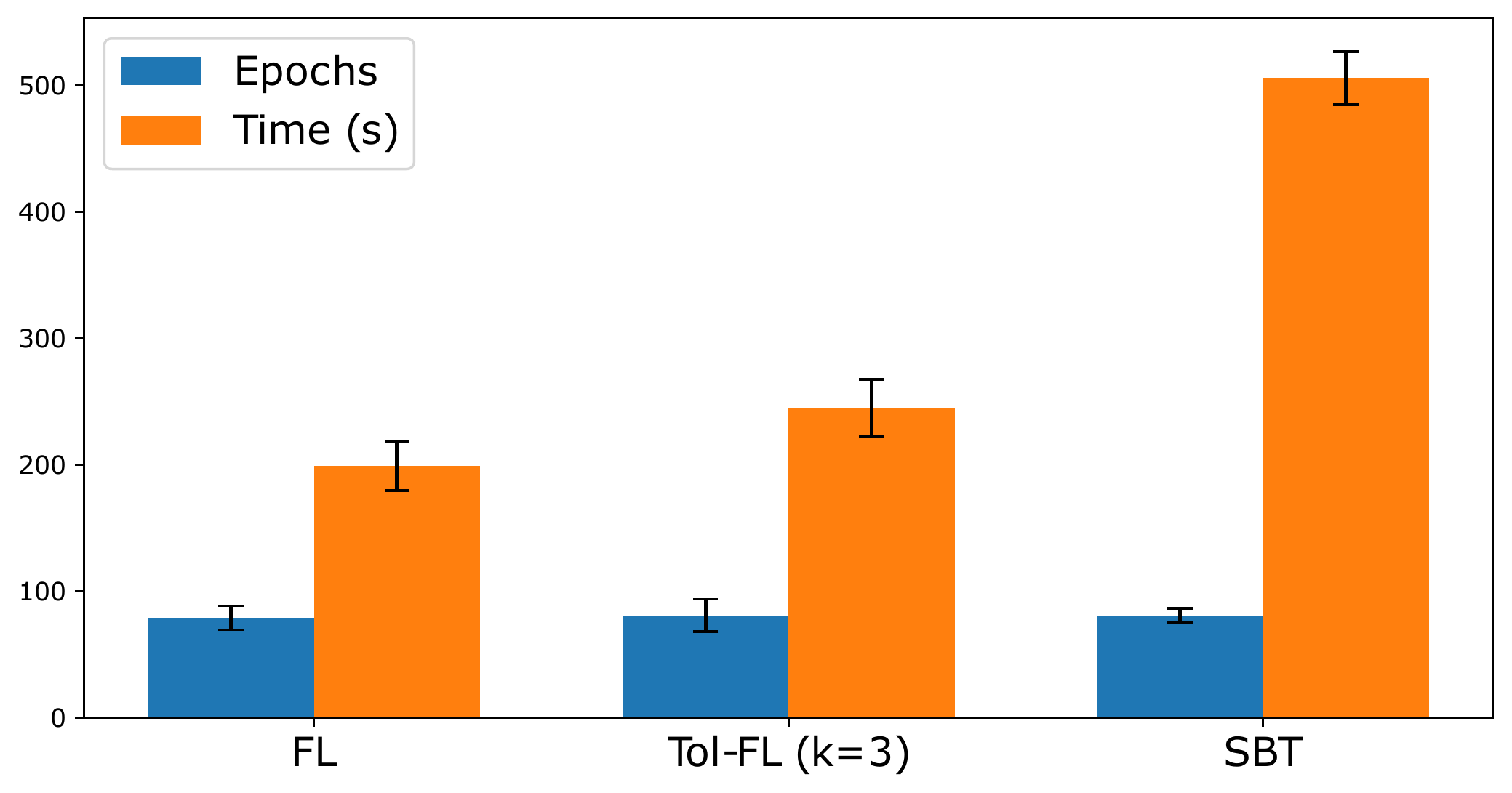}}
	\caption{Training time comparison---numbers of epochs and the wall-clock duration for each training method to reach the centralised (batch) training's converged loss (within 5\%, averaged over 5 trials).}
	\label{fig:timing_et}
\end{figure}

Regarding processing time, the training speed of Tol-FL, FL, and SBT (three opposing approaches that are expected to give the widest range) were compared under similar conditions to real-world applications.
This comparison was performed with each method being set to train an anomaly detection model for our wireless networking scenario until a predetermined training loss was reached.
The wall-clock time taken and number of epochs needed by each scheme to converge to the same loss as centralised (batch) training are shown in Figure~\ref{fig:timing_et}.
Compared to the theoretical estimates in Section~\ref{sec:training}, the trends in training time are as expected.
With FL and Tol-FL both allowing for parallel processing, these methods were shown to outperform centralised training in terms of time taken.
The differences between the distributed methods are attributable to the proportion of training that is carried out sequentially as opposed to in parallel, where Tol-FL is required to send aggregation messages to prime each device for parallel processing.

Regarding computational cost, the results of training for $100$ epochs in edge cases of Tol-FL (spanning FL to SBT) are shown in Table~\ref{tbl:comms_mb}.
The results show that we achieve a lower communication cost than competing methods.
This reduction is expected due to the greater proportion of communications that can occur within a cluster (without any feedback needed within each cluster) removing a substantial proportion of messages that occur in FL.
To arrive at the same low resource usage, a limit on communications cost in FL would require a restricted number of epochs, worsening performance.
Exploring this deterioration of performance is left to be addressed in future work.

\begin{table}
\caption{Distributed training communication costs}
\label{tbl:comms_mb}
\centering
\begin{tabular}{c|c|c}
\hline 
\textbf{Method} & \textbf{Expected} & \textbf{Actual (MB/epoch)} \\ 
\hline 
FL & $\mathcal{O}(2N)$ & $28.3$ \\ 
SBT & $\mathcal{O}(N)$ & $12.8$ \\ 
Tol-FL & $\mathcal{O}(N + k)$ & $21.0$ \\ 
\hline 
\end{tabular} 
\end{table}

% ====================
\section{Related Work}
Tol-FL is not the first method to build on FL.
However, as the following subsections will explain, Tol-FL addresses an oft-overlooked shortcoming in the standard FL definition, and explores possible improvements for the application of distributed anomaly detection.

\subsection{Existing Methods for Distributed Anomaly Detection}
Within a distributed anomaly detection problem, the task of identifying anomalies considers the collected samples at each distributed device, and defining what constitutes a typical sample.
For the case of network monitoring in an SDR context, these samples can be network readings from different geographic regions or segments of the frequency spectrum; any sample drawn from a different distribution from the training set should be identified as an anomaly.
The resulting problem is often viewed as a pattern recognition problem or a data representation task---learning what features are present in typical data samples \cite{rajendran_saife_2018, pfammatter_software-defined_2015, rajendran_electrosense_2018}.

The state-of-the-art literature for anomaly detection tasks consist of both geometric and ML approaches.
Previous studies have explored finding relevant features \cite{erfani_privacy-preserving_2014, lyu_improved_2016, lee_anomaly_2013, rajasegarar_ellipsoidal_2014} or training a ML-based anomaly detection model using distributed devices \cite{luo_distributed_2018, katzef_distributed_2020, srinu_physical_2019}.
Both FL and gossip-based techniques have been shown to approach theoretical performance limits \cite{lazarevic_theoretically_2009, bhuyan_network_2014}. Given these similarities in performance, a consideration of secondary features like resource usage and fragility becomes more important.

\subsection{Prior Federated Learning Research}
The FL training paradigm has been analysed \cite{li_convergence_2020, dinh_federated_2021}, adopted \cite{imteaj_survey_2021, peng_federated_2019, li_federated_nodate}, and improved upon in research and industry alike---seeing use in mainstream software like such as Google's Android Keyboard.
In such applications, the task of model training is moved to the devices at the edge of a network (such as smartphones) to collect local results through on-device training before aggregating multiple updates at a central location.
The underlying algorithm is FL's FedAvg, which utilises a star topology for computing which while computationally efficient, introduces risks relating to both device failure and data security, which recent works have attempted to rectify \cite{bonawitz_practical_2017, jiang_collaborative_2017, hegedus_gossip_2019}.

Existing studies have sought to improve the robustness of FL with new features that consider crash failures and Byzantine failures \cite{imteaj_survey_2021, wu_safa_2021} and support more adaptable schemes such as run-time task assignment \cite{stripelis_semi-synchronous_2021}.
These works, however, consider crashes of \textit{the client devices} instead of the server; the server is assumed to be infallible and active for the duration of an FL simulation.
These techniques ignore the realistic threat of server failure during model training, the failure that Tol-FL remedies through its use of a hybrid topology.

Previous works that have considered collaborative training and server failure include \cite{zhang_csafl_2021, duan_flexible_2021} in the context of FL and \cite{yang_scheduling_2019, hegedus_gossip_2019, ickin_privacy_2019} that consider gossip-based training schemes.
While gossip-based methods rely on random walks of data, the former, FL modification methods opt for a clustering method from which Tol-FL draws its inspiration.
Ignoring the constraints of local communication, the FL-based methods \cite{zhang_csafl_2021, duan_flexible_2021, chou_efficient_2021} determine a natural grouping scheme over all of the available devices based on the similarity between their datasets.
This modified scheme discards the speed benefit that arises through single-hop communications and practical considerations of link availability and instead forming virtual clusters that may include devices with large communications delays.
Other alternatives include hierarchical structures such as \cite{liu_client-edge-cloud_2020} which performs similar processing to Tol-FL but requires manual, time-consuming initial setup, and following a rigid schedule.

% ====================
\section{Conclusion}
Device failure can mean the difference between a smoothly-running collaborative system and a segregated set of devices left to perform sub-optimally.
Throughout this paper, we have explored the effect of client and server failures in the context of collaborative ML training, and proposed a novel training method---Tol-FL---that extends the de-facto standard FL so that a level of tolerance is introduced against server failure.
As FL does not feature any redundancy for its server, a method such as Tol-FL is needed for training to persist in the event of a server failure; continuing training with the existing clients without downtime, reconfiguration, or exposing any additional security risks.

Our work shows that in the event of server failure, Tol-FL outperforms FL in terms of model losses (and subsequently, anomaly detection), while maintaining similar performance in the case of client failure or standard, failure-free operation.
Furthermore, we show that Tol-FL achieves this increased failure tolerance while maintaining similar or reduced resource requirements compared to FL.
From this performance, Tol-FL shows an avenue forward for failure-tolerant FL to meet the needs that arise in volatile or hazardous environments in which neither clients nor server(s) are guaranteed to remain operational.

% ====================
\section*{Acknowledgment}
This work was supported in part by the Australian Research Council Linkage Project under the grant LP190101287, by Northrop Grumman Mission Systems' University Research Program, and by The University of Melbourne’s Research Computing Services and the Petascale Campus Initiative.

\appendix
\section{Appendix}

\subsection{Super-Batch Training}
\label{app:sbt}
A completely flat alternative to FL has been proposed without a name in prior work.
For simplicity, we dub this scheme ``Super-Batch Training" (SBT) due to its use of whole device datasets as if they were merely mini-batches.
In SBT, the global gradient, $g_{t}$, can be accurately obtained for model updates by iterating over the participating devices and combining their calculated gradients with the previous device's output, following a classical round robin scheme \cite{miao_zander_sung_ben_slimane_2016}.
This combination is a weighted sum between the local result and the mean that was calculated from previous samples, weighted based on sample sizes.
While batch training iterates over the entire dataset before updating weights, mini-batch training only considers a subset of the dataset.
In contrast, ``Super-Batch Training" (SBT) iterates over not just the full local dataset, but instead uses the full dataset across all devices before updating weights.
SBT updates a ML model only after passing over the data of multiple devices in a loop.

\begin{algorithm}[tb]
   \caption{One round of SBT, $SuperBatch(D, \theta_{t})$}
   \label{alg:superbatch}
	\begin{algorithmic}
	\REQUIRE $D, \theta_{t}$ // Devices and Model parameters at the current time step
	\STATE // On each device, \textbf{in parallel}:
	\FOR {i = 1, 2, ..., N}
		\STATE $g_{t, i} = \nabla_{\theta} J(X_i, \theta_{t})$ \algorithmiccomment{Calculate gradients}
	\ENDFOR

	// Initialise on the first device in a sequence
	\STATE $g_{t} = \textbf{0}$ // Mean gradient
	\STATE $n_{t} = 0$  // Sample count
	\STATE $N = |D|$ // The number of devices present
	\STATE // On each device, \textbf{in sequence}:
	\FOR {i = 1, 2, ..., N}
		\STATE $n_{t} \leftarrow n_{t} + | X_i |$  // Record the number of local samples
		\STATE $r \leftarrow \frac{| X_i |}{n_{t}}$  // Calculate a weighting factor
		\STATE $g_{t} \leftarrow r g_{t, i} + (1 - r) g_{t} $ // Update the mean gradient
	\ENDFOR
	\STATE // On the last device of the sequence, apply gradient update
	\STATE $\theta_{t+1} = \theta_{t} - \alpha g_{t}$
	\STATE // Broadcast the latest parameters, $\theta_{t+1}$
	\end{algorithmic}
\end{algorithm}

Algorithm~\ref{alg:superbatch} shows how SBT is performed, using the same steps of batch training but with additional communication between devices.
In this algorithm, the task given to each participating device is simply to calculate training gradients on their local dataset with the model parameters $\theta_t$. After calculating local gradients, these gradients are sequentially exchanged with peers in a communication-efficient and flexible scheme.
By delaying all parameter updates until \textit{all} selected devices have been traversed, all gradients are calculated for \textit{the same} model parameters. Local gradients can then be seen as equivalent to those constructed using batch training on a single device.

\begin{figure}[h]
	\centerline{\includegraphics[width=0.3\textwidth]{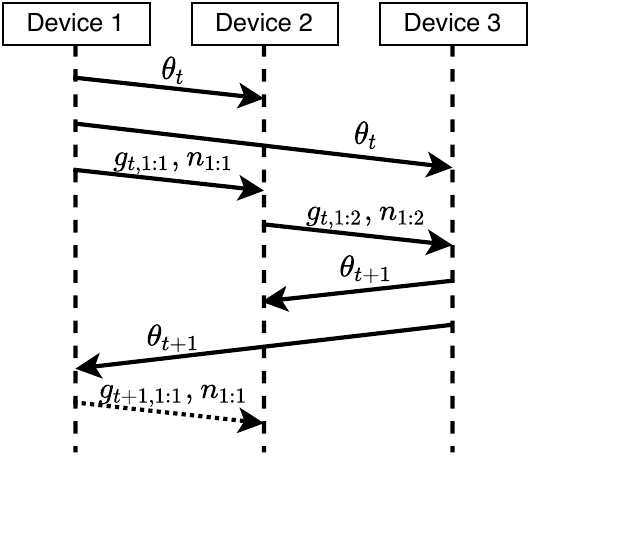}}
	\caption{SBT sequence diagram showing all communication steps involved in one global model update from $\theta_{t}$ to $\theta_{t+1}$}
	\label{fig:sb_seq}
\end{figure}

While SBT mirrors the weighting scheme of FL, its flat architecture introduces two key differences: the lack of need of a server; and the increase in communication time spent transferring models in sequence instead of in parallel, highlighted in Figure~\ref{fig:sb_seq}.
The inherent trade off in these properties makes the two schemes more appropriate for different network topologies, with SBT more suited to mesh networks and structures where there are concerns regarding the robustness of the centralised server.
This increased failure tolerance introduces an overhead in terms of the scheduling and coordination of the gradient updates, due to their sequential nature.
In contrast, FL is more suited to star-topologies.

Super-Batch Training performs model updates that are identical to centralised, batch training.
From first principles, using $\hat{f}_{a,b}$ to represent the mean of samples $f_a$ through $f_{b-1}$ or
\begin{equation}
\hat{f}_{a, b} = \underset{i = a, ..., b-1}{\mathbb{E}}[f_i] = \frac{1}{b - a} \sum_{i = a}^{b-1} f_i,
\end{equation}
which allows the global mean to be iteratively calculated by
\begin{equation}
\hat{f}_{a, c} = \frac{1}{c - a} \left( (b - a)\hat{f}_{a, b} + (c - b)\hat{f}_{b, c} \right)
\end{equation}
until the entire dataset is covered. Using the global gradient $g_t$ as a function that depends on devices $1$ through $n_t$ $\hat{f}_{1, n_t}$, each new device extends the range of samples included in the mean ($n_{t} \rightarrow n_{t+1}$), and efficiently adjust the mean ($g_t$) to account for the new entries.

\subsection{Reference Datasets}
\label{app:datasets}
Each of the datasets described in Table~\ref{tbl:data} was cast as an anomaly detection task by designating one or more classes within each dataset as ``anomalous" data.
The remaining data classes were divided amongst the client devices in a simulation. Where clusters are present, this data was assigned as one class per cluster.

\begin{table}[htp]
\centering
\caption{The datasets used to test Tol-FL training.}
% sample size, data count, anomaly count, typical data
\begin{tabular}{c|c|c|c}
\hline
%\hline
Dataset & Sample shape & \# Classes & \# Samples/class \\
\hline
Fashion-MNIST & $28 \times 28$ & $10$ & $7,000$\\
CIFAR-10 & $32 \times 32$ & $10$ & $7,000$\\
CIFAR-100 & $32 \times 32$ & $100$ & $500$\\
Comms-ML & $112 \times 1$ & $4$ & $3,000$ \\
\hline
\end{tabular}
\label{tbl:data}
\end{table}

% Can use something like this to put references on a page
% by themselves when using endfloat and the captionsoff option.
\ifCLASSOPTIONcaptionsoff
  \newpage
\fi

\bibliographystyle{IEEEtran}
% argument is your BibTeX string definitions and bibliography database(s)
\bibliography{IEEEabrv,references/papers}

% Generated by IEEEtran.bst, version: 1.14 (2015/08/26)
\begin{thebibliography}{10}
\providecommand{\url}[1]{#1}
\csname url@samestyle\endcsname
\providecommand{\newblock}{\relax}
\providecommand{\bibinfo}[2]{#2}
\providecommand{\BIBentrySTDinterwordspacing}{\spaceskip=0pt\relax}
\providecommand{\BIBentryALTinterwordstretchfactor}{4}
\providecommand{\BIBentryALTinterwordspacing}{\spaceskip=\fontdimen2\font plus
\BIBentryALTinterwordstretchfactor\fontdimen3\font minus
  \fontdimen4\font\relax}
\providecommand{\BIBforeignlanguage}[2]{{%
\expandafter\ifx\csname l@#1\endcsname\relax
\typeout{** WARNING: IEEEtran.bst: No hyphenation pattern has been}%
\typeout{** loaded for the language `#1'. Using the pattern for}%
\typeout{** the default language instead.}%
\else
\language=\csname l@#1\endcsname
\fi
#2}}
\providecommand{\BIBdecl}{\relax}
\BIBdecl

\bibitem{zhang_deep_2018}
C.~Zhang, P.~Patras, and H.~Haddadi, ``Deep learning in mobile and wireless
  networking: A survey,'' \emph{IEEE Communications Surveys Tutorials},
  vol.~21, no.~3, pp. 2224--2287, 2019.

\bibitem{liu_survey_2017}
W.~Liu, Z.~Wang, X.~Liu, N.~Zeng, Y.~Liu, and F.~E. Alsaadi,
  ``\BIBforeignlanguage{en}{A survey of deep neural network architectures and
  their applications},'' \emph{\BIBforeignlanguage{en}{Neurocomputing}}, vol.
  234, pp. 11--26, Apr. 2017.

\bibitem{jiang_collaborative_2017}
Z.~Jiang, A.~Balu, C.~Hegde, and S.~Sarkar, ``Collaborative {Deep} {Learning}
  in {Fixed} {Topology} {Networks},'' \emph{arXiv:1706.07880 [cs, stat]}, Jun.
  2017, arXiv: 1706.07880.

\bibitem{mcmahan2017communication}
B.~McMahan, E.~Moore, D.~Ramage, S.~Hampson, and B.~A. y~Arcas,
  ``Communication-{Efficient} {Learning} of {Deep} {Networks} from
  {Decentralized} {Data},'' in \emph{Artificial {I}ntelligence and
  {S}tatistics}.\hskip 1em plus 0.5em minus 0.4em\relax PMLR, 2017, pp.
  1273--1282.

\bibitem{bonawitz_practical_2017}
\BIBentryALTinterwordspacing
K.~Bonawitz, V.~Ivanov, B.~Kreuter, A.~Marcedone, H.~B. McMahan, S.~Patel,
  D.~Ramage, A.~Segal, and K.~Seth, ``\BIBforeignlanguage{en}{Practical
  {Secure} {Aggregation} for {Privacy}-{Preserving} {Machine} {Learning}},'' in
  \emph{\BIBforeignlanguage{en}{Proceedings of the 2017 {ACM} {SIGSAC}
  {Conference} on {Computer} and {Communications} {Security}}}.\hskip 1em plus
  0.5em minus 0.4em\relax Dallas Texas USA: ACM, Oct. 2017, pp. 1175--1191.
  [Online]. Available: \url{https://dl.acm.org/doi/10.1145/3133956.3133982}
\BIBentrySTDinterwordspacing

\bibitem{dinh_federated_2021}
\BIBentryALTinterwordspacing
C.~T. Dinh, N.~H. Tran, M.~N.~H. Nguyen, C.~S. Hong, W.~Bao, A.~Y. Zomaya, and
  V.~Gramoli, ``Federated {Learning} over {Wireless} {Networks}: {Convergence}
  {Analysis} and {Resource} {Allocation},'' \emph{IEEE/ACM Transactions on
  Networking}, vol.~29, no.~1, pp. 398--409, Feb. 2021, number: 1 arXiv:
  1910.13067. [Online]. Available: \url{http://arxiv.org/abs/1910.13067}
\BIBentrySTDinterwordspacing

\bibitem{kwon_survey_2019}
D.~Kwon, H.~Kim, J.~Kim, S.~C. Suh, I.~Kim, and K.~J. Kim, ``A survey of deep
  learning-based network anomaly detection,'' \emph{Cluster Computing},
  vol.~22, no.~1, pp. 949--961, Jan 2019.

\bibitem{duan_fedgroup_2021}
\BIBentryALTinterwordspacing
M.~Duan, D.~Liu, X.~Ji, R.~Liu, L.~Liang, X.~Chen, and Y.~Tan,
  ``\BIBforeignlanguage{en}{{FedGroup}: {Efficient} {Clustered} {Federated}
  {Learning} via {Decomposed} {Data}-{Driven} {Measure}},'' Jul. 2021, number:
  arXiv:2010.06870 arXiv:2010.06870 [cs]. [Online]. Available:
  \url{http://arxiv.org/abs/2010.06870}
\BIBentrySTDinterwordspacing

\bibitem{NEURIPS2020_e32cc80b}
A.~Ghosh, J.~Chung, D.~Yin, and K.~Ramchandran, ``An efficient framework for
  clustered federated learning,'' in \emph{Advances in Neural Information
  Processing Systems}, H.~Larochelle, M.~Ranzato, R.~Hadsell, M.~Balcan, and
  H.~Lin, Eds., vol.~33.\hskip 1em plus 0.5em minus 0.4em\relax Curran
  Associates, Inc., 2020, pp. 19\,586--19\,597.

\bibitem{xie_multi-center_2021}
\BIBentryALTinterwordspacing
M.~Xie, G.~Long, T.~Shen, T.~Zhou, X.~Wang, J.~Jiang, and C.~Zhang,
  ``\BIBforeignlanguage{en}{Multi-{Center} {Federated} {Learning}},'' Aug.
  2021, number: arXiv:2005.01026 arXiv:2005.01026 [cs, stat]. [Online].
  Available: \url{http://arxiv.org/abs/2005.01026}
\BIBentrySTDinterwordspacing

\bibitem{chou_efficient_2021}
L.~Chou, Z.~Liu, Z.~Wang, and A.~Shrivastava, ``Efficient and {Less}
  {Centralized} {Federated} {Learning},'' in \emph{{Joint} {European}
  {Conference} on {Machine} {Learning} and {Knowledge} {Discovery} in
  {Databases}}.\hskip 1em plus 0.5em minus 0.4em\relax Springer, 2021, pp.
  772--787.

\bibitem{yang_scheduling_2019}
H.~H. Yang, Z.~Liu, T.~Q.~S. Quek, and H.~V. Poor, ``Scheduling {Policies} for
  {Federated} {Learning} in {Wireless} {Networks},'' \emph{arXiv:1908.06287
  [cs, eess, math]}, Oct. 2019.

\bibitem{goodfellow_deep_2016}
I.~Goodfellow, Y.~Bengio, and A.~Courville, \emph{Deep {Learning}}.\hskip 1em
  plus 0.5em minus 0.4em\relax MIT Press, 2016.

\bibitem{goodfellow_empirical_2015}
I.~J. Goodfellow, M.~Mirza, D.~Xiao, A.~Courville, and Y.~Bengio, ``An
  {Empirical} {Investigation} of {Catastrophic} {Forgetting} in
  {Gradient}-{Based} {Neural} {Networks},'' \emph{arXiv:1312.6211 [cs, stat]},
  Mar. 2015, arXiv: 1312.6211.

\bibitem{katzef_commsml}
\BIBentryALTinterwordspacing
M.~Katzef, ``Comms-ml: Communications ml dataset generator,'' 2022. [Online].
  Available: \url{https://github.com/mkatzef/comms-ml}
\BIBentrySTDinterwordspacing

\bibitem{rajendran_saife_2018}
S.~Rajendran, W.~Meert, V.~Lenders, and S.~Pollin, ``{SAIFE}: {Unsupervised}
  {Wireless} {Spectrum} {Anomaly} {Detection} with {Interpretable}
  {Features},'' in \emph{2018 {IEEE} {International} {Symposium} on {Dynamic}
  {Spectrum} {Access} {Networks} ({DySPAN})}, Oct. 2018, pp. 1--9, iSSN:
  2334-3125.

\bibitem{pfammatter_software-defined_2015}
D.~Pfammatter, D.~Giustiniano, and V.~Lenders, ``\BIBforeignlanguage{en}{A
  software-defined sensor architecture for large-scale wideband spectrum
  monitoring},'' in \emph{\BIBforeignlanguage{en}{The 14th {International}
  {Conference} on {Information} {Processing} in {Sensor} {Networks} - {IPSN}
  '15}}.\hskip 1em plus 0.5em minus 0.4em\relax Seattle, Washington: ACM Press,
  2015, pp. 71--82.

\bibitem{rajendran_electrosense_2018}
S.~Rajendran, R.~Calvo-Palomino, M.~Fuchs, B.~V.~d. Bergh, H.~Cordobés,
  D.~Giustiniano, S.~Pollin, and V.~Lenders, ``Electrosense: {Open} and {Big}
  {Spectrum} {Data},'' \emph{IEEE Communications Magazine}, vol.~56, no.~1, pp.
  210--217, Jan. 2018, arXiv: 1703.09989.

\bibitem{erfani_privacy-preserving_2014}
S.~M. Erfani, Y.~W. Law, S.~Karunasekera, C.~A. Leckie, and M.~Palaniswami,
  ``\BIBforeignlanguage{en}{Privacy-{Preserving} {Collaborative} {Anomaly}
  {Detection} for {Participatory} {Sensing}},'' in
  \emph{\BIBforeignlanguage{en}{Advances in {Knowledge} {Discovery} and {Data}
  {Mining}}}, ser. Lecture {Notes} in {Computer} {Science}, V.~S. Tseng, T.~B.
  Ho, Z.-H. Zhou, A.~L.~P. Chen, and H.-Y. Kao, Eds.\hskip 1em plus 0.5em minus
  0.4em\relax Cham: Springer International Publishing, 2014, pp. 581--593.

\bibitem{lyu_improved_2016}
L.~Lyu, Y.~W. Law, S.~M. Erfani, C.~Leckie, and M.~Palaniswami,
  ``\BIBforeignlanguage{en}{An improved scheme for privacy-preserving
  collaborative anomaly detection},'' in \emph{\BIBforeignlanguage{en}{2016
  {IEEE} {International} {Conference} on {Pervasive} {Computing} and
  {Communication} {Workshops} ({PerCom} {Workshops})}}.\hskip 1em plus 0.5em
  minus 0.4em\relax Sydney, Australia: IEEE, Mar. 2016, pp. 1--6.

\bibitem{lee_anomaly_2013}
Y.-J. Lee, Y.-R. Yeh, and Y.-C.~F. Wang, ``\BIBforeignlanguage{en}{Anomaly
  {Detection} via {Online} {Oversampling} {Principal} {Component}
  {Analysis}},'' \emph{\BIBforeignlanguage{en}{IEEE Transactions on Knowledge
  and Data Engineering}}, vol.~25, no.~7, pp. 1460--1470, Jul. 2013.

\bibitem{rajasegarar_ellipsoidal_2014}
S.~Rajasegarar, A.~Gluhak, M.~Ali~Imran, M.~Nati, M.~Moshtaghi, C.~Leckie, and
  M.~Palaniswami, ``\BIBforeignlanguage{en}{Ellipsoidal neighbourhood outlier
  factor for distributed anomaly detection in resource constrained networks},''
  \emph{\BIBforeignlanguage{en}{Pattern Recognition}}, vol.~47, no.~9, pp.
  2867--2879, Sep. 2014.

\bibitem{luo_distributed_2018}
T.~Luo and S.~G. Nagarajan, ``\BIBforeignlanguage{en}{Distributed {Anomaly}
  {Detection} {Using} {Autoencoder} {Neural} {Networks} in {WSN} for {IoT}},''
  in \emph{\BIBforeignlanguage{en}{2018 {IEEE} {International} {Conference} on
  {Communications} ({ICC})}}.\hskip 1em plus 0.5em minus 0.4em\relax Kansas
  City, MO: IEEE, May 2018, pp. 1--6.

\bibitem{katzef_distributed_2020}
M.~Katzef, A.~C. Cullen, T.~Alpcan, C.~Leckie, and J.~Kopacz,
  ``\BIBforeignlanguage{en}{Distributed {Generative} {Adversarial} {Networks}
  for {Anomaly} {Detection}},'' in \emph{\BIBforeignlanguage{en}{Decision and
  {Game} {Theory} for {Security}}}, ser. Lecture {Notes} in {Computer}
  {Science}, Q.~Zhu, J.~S. Baras, R.~Poovendran, and J.~Chen, Eds.\hskip 1em
  plus 0.5em minus 0.4em\relax Cham: Springer International Publishing, 2020,
  pp. 3--22.

\bibitem{srinu_physical_2019}
S.~Srinu, M.~K.~K. Reddy, and C.~Temaneh-Nyah, ``Physical layer security
  against cooperative anomaly attack using bivariate data in distributed
  {CRNs},'' in \emph{2019 11th {International} {Conference} on {Communication}
  {Systems} {Networks} ({COMSNETS})}, Jan. 2019, pp. 410--413, iSSN: 2155-2509.

\bibitem{lazarevic_theoretically_2009}
A.~Lazarevic, N.~Srivastava, A.~Tiwari, J.~Isom, N.~Oza, and J.~Srivastava,
  ``\BIBforeignlanguage{en}{Theoretically {Optimal} {Distributed} {Anomaly}
  {Detection}},'' in \emph{\BIBforeignlanguage{en}{2009 {IEEE} {International}
  {Conference} on {Data} {Mining} {Workshops}}}.\hskip 1em plus 0.5em minus
  0.4em\relax Miami, FL, USA: IEEE, Dec. 2009, pp. 515--520.

\bibitem{bhuyan_network_2014}
M.~H. Bhuyan, D.~K. Bhattacharyya, and J.~K. Kalita,
  ``\BIBforeignlanguage{en}{Network {Anomaly} {Detection}: {Methods}, {Systems}
  and {Tools}},'' \emph{\BIBforeignlanguage{en}{IEEE Communications Surveys \&
  Tutorials}}, vol.~16, no.~1, pp. 303--336, 2014.

\bibitem{li_convergence_2020}
\BIBentryALTinterwordspacing
X.~Li, K.~Huang, W.~Yang, S.~Wang, and Z.~Zhang, ``On the {Convergence} of
  {FedAvg} on {Non}-{IID} {Data},'' \emph{arXiv:1907.02189 [cs, math, stat]},
  Jun. 2020, arXiv: 1907.02189. [Online]. Available:
  \url{http://arxiv.org/abs/1907.02189}
\BIBentrySTDinterwordspacing

\bibitem{imteaj_survey_2021}
A.~Imteaj, U.~Thakker, S.~Wang, J.~Li, and M.~H. Amini, ``A {Survey} on
  {Federated} {Learning} for {Resource}-{Constrained} {IoT} {Devices},''
  \emph{IEEE Internet of Things Journal}, pp. 1--1, 2021, conference Name: IEEE
  Internet of Things Journal.

\bibitem{peng_federated_2019}
X.~Peng, Z.~Huang, Y.~Zhu, and K.~Saenko, ``Federated {Adversarial} {Domain}
  {Adaptation},'' \emph{arXiv:1911.02054 [cs]}, Dec. 2019, arXiv: 1911.02054.

\bibitem{li_federated_nodate}
Y.~Li, ``\BIBforeignlanguage{en}{Federated {Learning} for {Time} {Series}
  {Forecasting} {Using} {Hybrid} {Model}},'' in
  \emph{\BIBforeignlanguage{en}{Degree Project Thesis, KTH Royal Institute of
  Technology, Sweden, 2019}}, p.~76.

\bibitem{hegedus_gossip_2019}
I.~Hegedűs, G.~Danner, and M.~Jelasity, ``\BIBforeignlanguage{en}{Gossip
  {Learning} as a {Decentralized} {Alternative} to {Federated} {Learning}},''
  in \emph{\BIBforeignlanguage{en}{Distributed {Applications} and
  {Interoperable} {Systems}}}, ser. Lecture {Notes} in {Computer} {Science},
  J.~Pereira and L.~Ricci, Eds.\hskip 1em plus 0.5em minus 0.4em\relax Cham:
  Springer International Publishing, 2019, pp. 74--90.

\bibitem{wu_safa_2021}
W.~Wu, L.~He, W.~Lin, R.~Mao, C.~Maple, and S.~Jarvis, ``{SAFA}: {A}
  {Semi}-{Asynchronous} {Protocol} for {Fast} {Federated} {Learning} {With}
  {Low} {Overhead},'' \emph{IEEE Transactions on Computers}, vol.~70, no.~5,
  pp. 655--668, May 2021, conference Name: IEEE Transactions on Computers.

\bibitem{stripelis_semi-synchronous_2021}
\BIBentryALTinterwordspacing
D.~Stripelis and J.~L. Ambite, ``Semi-{Synchronous} {Federated} {Learning},''
  \emph{arXiv:2102.02849 [cs]}, Feb. 2021, arXiv: 2102.02849. [Online].
  Available: \url{http://arxiv.org/abs/2102.02849}
\BIBentrySTDinterwordspacing

\bibitem{zhang_csafl_2021}
\BIBentryALTinterwordspacing
Y.~Zhang, M.~Duan, D.~Liu, L.~Li, A.~Ren, X.~Chen, Y.~Tan, and C.~Wang,
  ``{CSAFL}: {A} {Clustered} {Semi}-{Asynchronous} {Federated} {Learning}
  {Framework},'' \emph{arXiv:2104.08184 [cs]}, Apr. 2021, arXiv: 2104.08184.
  [Online]. Available: \url{http://arxiv.org/abs/2104.08184}
\BIBentrySTDinterwordspacing

\bibitem{duan_flexible_2021}
\BIBentryALTinterwordspacing
M.~Duan, D.~Liu, X.~Ji, Y.~Wu, L.~Liang, X.~Chen, and Y.~Tan, ``Flexible
  {Clustered} {Federated} {Learning} for {Client}-{Level} {Data} {Distribution}
  {Shift},'' \emph{arXiv:2108.09749 [cs]}, Aug. 2021, arXiv: 2108.09749.
  [Online]. Available: \url{http://arxiv.org/abs/2108.09749}
\BIBentrySTDinterwordspacing

\bibitem{ickin_privacy_2019}
S.~Ickin, K.~Vandikas, and M.~Fiedler, ``Privacy preserving qoe modeling using
  collaborative learning,'' in \emph{Proceedings of the 4th Internet-QoE
  Workshop on QoE-Based Analysis and Management of Data Communication
  Networks}.\hskip 1em plus 0.5em minus 0.4em\relax New York, NY, USA:
  Association for Computing Machinery, 2019, pp. 13--18.

\bibitem{liu_client-edge-cloud_2020}
L.~Liu, J.~Zhang, S.~Song, and K.~B. Letaief, ``Client-{Edge}-{Cloud}
  {Hierarchical} {Federated} {Learning},'' in \emph{{ICC} 2020 - 2020 {IEEE}
  {International} {Conference} on {Communications} ({ICC})}, Jun. 2020, pp.
  1--6, iSSN: 1938-1883.

\bibitem{miao_zander_sung_ben_slimane_2016}
G.~Miao, J.~Zander, K.~W. Sung, and S.~Ben~Slimane, \emph{Fundamentals of
  Mobile Data Networks}.\hskip 1em plus 0.5em minus 0.4em\relax Cambridge
  University Press, 2016.

\end{thebibliography}

\end{document}